\documentclass[conference]{IEEEtran} 
\ifCLASSINFOpdf
  \usepackage[pdftex]{graphicx}
\else
\fi
\usepackage{cite}
\usepackage{array}
\usepackage{booktabs}
\usepackage{tikz}
\usetikzlibrary{shapes.geometric, arrows}
\tikzstyle{io} = [trapezium, trapezium left angle=70, trapezium right angle=110]

\usepackage{smartdiagram}
\usesmartdiagramlibrary{additions}
\usepackage{hhline}
\usepackage{multirow}
\usepackage{amsmath}
\usepackage[nomessages]{fp} 
\usepackage{dblfloatfix} 

\begin{document}
%
\title{Synergistic Union of Word2Vec and Lexicon for Domain Specific Semantic Similarity}

\author{\IEEEauthorblockN{Keet Sugathadasa\IEEEauthorrefmark{1}, Buddhi Ayesha\IEEEauthorrefmark{1}, Nisansa de Silva\IEEEauthorrefmark{1}, Amal Shehan Perera\IEEEauthorrefmark{1},\\Vindula Jayawardana\IEEEauthorrefmark{1}, Dimuthu Lakmal\IEEEauthorrefmark{1}, Madhavi Perera\IEEEauthorrefmark{2}}
\IEEEauthorblockA{\IEEEauthorrefmark{1}Department of Computer Science \& Engineering\\University of Moratuwa}
\IEEEauthorblockA{\IEEEauthorrefmark{2}University of London International Programmes\\University of London}
\IEEEauthorblockA{Email: keetmalin.13@cse.mrt.ac.lk}
}


%


\maketitle

\begin{abstract}
Semantic similarity measures are an important part in Natural Language Processing tasks. However Semantic similarity measures built for general use do not perform well within specific domains. Therefore in this study we introduce a domain specific semantic similarity measure that was created by the synergistic union of word2vec, a word embedding method that is used for semantic similarity calculation and lexicon based (lexical) semantic similarity methods. We prove that this proposed methodology out performs word embedding methods trained on generic corpus and methods trained on domain specific corpus but do not use lexical semantic similarity methods to augment the results. Further, we prove that text lemmatization can improve the performance of word embedding methods.\\       

\end{abstract}


%
\IEEEpeerreviewmaketitle


\newcommand{\wvG}{$word2vec_{\scriptscriptstyle G}$}
\newcommand{\wvLR}{$word2vec_{\scriptscriptstyle LR}$}
\newcommand{\wvLL}{$word2vec_{\scriptscriptstyle LL}$}
\newcommand{\wvLLS}{$word2vec_{\scriptscriptstyle LLS}$}

\FPeval{\CaseCount}{clip(35000)} 
\FPeval{\WordBillionCount}{clip(20)} 
\FPeval{\GoldenConceptCount}{clip(100)} 
\FPeval{\GoldenLength}{clip(5)} 
\FPeval{\GoldenChoices}{clip(15)} 

\FPeval{\GoldenExtraCount}{clip(\GoldenConceptCount*\GoldenLength)} 
\FPeval{\GoldenPool}{clip(\GoldenConceptCount*\GoldenChoices)} 


\noindent\textbf{\textit{keywords}}: Word Embedding, Semantic Similarity, Neural Networks, Lexicon, word2vec

\section{Introduction}

Semantic Similarity measurements based on linguistic features are a fundamental component of almost all Natural Language Processing (NLP) tasks: Information Retrieval, Information Extraction, and Natural Language Understanding~\cite{wimalasuriya2010ontology}. In the case of NLP based Information Retrieval (IR), it plays into the task of obtaining the items that are most relevant to the query whereas Information Extraction (IE), plays into the task of correctly recognizing the linguistic elements to be extracted be it the Part of Speech (PoS) tags or Named Entities in Named Entity Recognition (NER). In the case of Text Understanding which is also known as Natural Language Understanding (NLU), it helps in identifying semantic connections between elements on the document that is being analyzed. 


Law and order could be rather regarded as the cloak of invisibility that operates and controls the human behavior to its possible extents in the name of justice. Thus in terms of maintaining social order, quiddity of law within the society is mandatory. John Stuart Mill articulated a principle in On Liberty, where he stated that \textit{The only purpose for which power can be rightfully exercised over any member of a civilized community, against his will, is to prevent harm to others}~\cite{mill1999liberty}. Being such a vital field, acquisition of laws and legal documents through technological means is on the list of necessities on the rise. Also, ample justification for the need of semantic disambiguation in the legal domain can be found in seminal cases such as: \textit{Fagan v MPC} and \textit{R v Woollin}. Therefore we selected the law as the domain for this study.

The legal domain contains a considerable amount of domain specific jargon where the etymology lies with mainly Latin and English. To complicate this fact, in certain cases, the meaning of the words and context differs by the legal officers interpretations. 
	
Another field which suffers similarly from this issue is the medical industry~\cite{oliver1999representation}. Non-systematic organization and complexity of the medical documents result in medical domain falling victim to loss of having proper representations for its terminology. PubMed~\cite{PubMed} attempts to remedy this. Later studies such as~\cite{de2017Discovering} utilize these repositories. Therefore, it is possible to claim that the problem addressed in this study is one that is not just limited to the legal domain but is one that transcends into a numerous other domains as well.


Methods that treat words as independent atomic units is not sufficient to capture the expressiveness of language~\cite{mikolov2013efficient}. A solution to this is word context learning methods~\cite{bengio2003neural,de2011semap,ng1996integrating}. Another solution is lexical semantic similarity based methods~\cite{deSilva2015safs3}. Both of these approaches try to capture semantic and syntactic information of a word. In this study we propose a methodology to have a synergistic union of both of these methods. First for word context learning, we used a Word Embedding~\cite{bengio2003neural} method, word2vec~\cite{mikolov2013efficient}. Then we used a number of lexical semantic similarity measures~\cite{Wu1994,Jiang1997,hirst1998lexical} to augment and improve the result.




The hypothesis of this study has three main claims: (1) A word embedding model trained on a small domain specific corpus can outperform a word embedding model trained on a large but generic corpus, (2) Word lemmatization, which removes inflected forms of words, would improve the performance of a word embedding model, (3) Usage of lexical semantic similarity measures trained over a machine learning system can improve the overall system performance. Our results sufficiently prove all of these claims to be true.

The structure of the paper is as follows. Section~\ref{Background} gives a brief overview on the current tools being used and domains that have tackled this problem of word representations. Section~\ref{Methodology} gives a description on the methodology being used in this research in order to obtain the results and conclusions as necessary. That is followed by Section~\ref{Results} that presents and analyses results. The paper ends with Section~\ref{conclusion} which gives the conclusion and discusses future work.

\section{Background and Related Work}
\label{Background}
This section illustrates the background of the techniques used in this study and work carried out by others in various areas relevant to this research. The subsections given below are the important key areas in this study.

\subsection{Lexical Semantic Similarity Measures}
Lexical Semantic similarity of two entities is a measure of the likeness of the semantic content of those entities, most commonly calculated with the help of topological similarity existing within an ontology such as WordNet~\cite{miller1990introduction}. Wu and Palmer proposed a method to give the similarity between two words in the 0 to 1 range~\cite{Wu1994}. In comparison, Jiang and Conrath proposed a method to measure the lexical semantic similarity between word pairs using corpus statistics and lexical taxonomy~\cite{Jiang1997}. Hirst \& St-Onge's system~\cite{hirst1998lexical} quantifies the amount that the relevant synsets are connected by a path that is not too long and that does not change direction often. The strengths of each of these algorithms were evaluated in~\cite{deSilva2015safs3} by means of~\cite{WS4J}. 

\subsection{Word Vector Embedding}

Traditionally, in Natural Language Processing systems, words are treated as atomic units ignoring the correlation between the words as they are represented just by indices in a vocabulary~\cite{mikolov2013efficient}. To solve the inadequacies of that approach, distributed representation of words and phrases through word embeddings was proposed~\cite{mikolov2013distributed}. The idea is to create vector representations for each of the words in a text document along with word meanings and relationships between the words all mapped to a common vector space.

A number of Word Vector Embedding systems have been proposed such as: GloVe~\cite{pennington2014glove}, Latent Dirichlet Allocation (LDA)~\cite{das2015gaussian}, and word2vec\footnote{https://code.google.com/p/word2vec/}\cite{goldberg2014word2vec}. GloVe uses a \textit{word to neighboring word} mapping when learning dense embeddings, which uses a matrix factorization mechanism. LDA uses a similar approach via matrices, but the concept is based on mapping words with relevant sets of documents. word2vec uses a neural network based approach that uses \textit{word to neighboring word} mapping. Due to the flexibility and features it provided in terms of parameter passing when training the model using a text corpus, we use word2vec in this study. word2vec supports two main training models: Skip-gram~\cite{melamud2015simple} and Continuous Bag Of Words (CBOW)~\cite{goldberg2014word2vec}.

\subsection{Legal Information Systems}

Schweighofer~\cite{schweighofer1993legal}, claims that there is a huge vacuum that should be addressed in eradicating the information crisis that the applications in the field of law suffer from. This vacuum is evident by the fact that, despite being important, there is a scarcity of legal information systems. Even though the two main commercial systems; WestLaw\footnote{https://www.westlaw.com/} and LexisNexis\footnote{https://www.lexisnexis.com/} are widely used, they only provide query based searching, where legal officers need to remember keywords which are predefined when querying for relevant legal information, there is still a hassle in accessing this information.  

One of the most popular legal information retrieval systems is KONTERM~\cite{schweighofer1993legal}, which was developed to represent document structures and contents. However, it too suffered from scalability issues. The currently existing implementation that is closest to our proposed model is Gov2Vec~\cite{nay2016gov2vec}, which is a system that creates vector representations of words in the legal domain, by creating a vocabulary from across all corpora on, supreme court opinions, presidential actions and official summaries of congressional bills. It uses a neural network~\cite{bengio2003neural} to predict the target word with the mean of its context words' vectors. However, the text copora used here, itself was not sufficient enough to represent the entire legal domain. In addition to that, the Gov2Vec trained model is not available to use used by legal professionals or to be tested against. 





\section{Methodology}
\label{Methodology}

\begin{figure*}[!htb]	
	\centering
\includegraphics[width=0.92\textwidth]{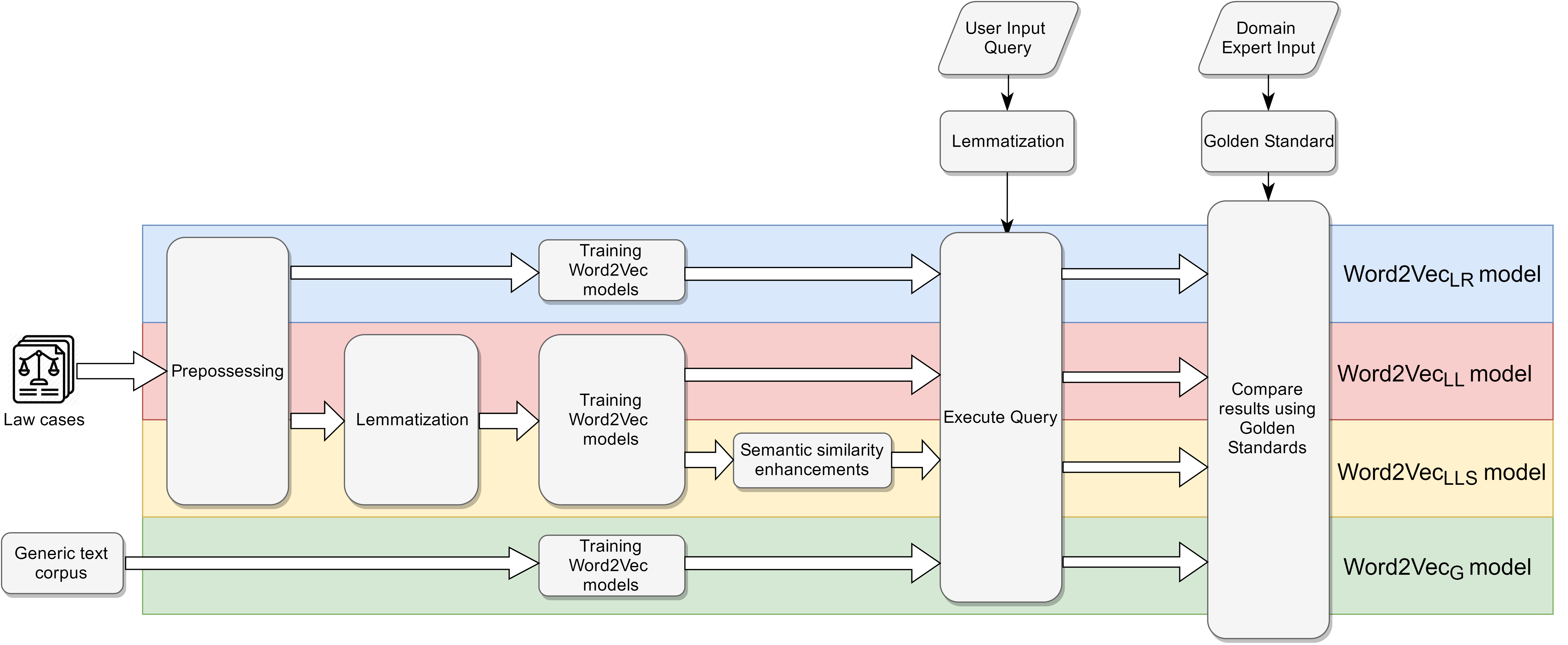}
	\caption{Flow diagram for the Overall Methodology}	
	\label{fig:Overall methodology}
\end{figure*}

This section describes the research that was carried out. Each section below, addresses a component in the overall methodology. An overview of the methodology we propose is illustrated in~\figurename~\ref{fig:Overall methodology}.

The first phase of this study was to gather the necessary legal cases from on-line repositories. We obtained over $\CaseCount$ legal case documents, pertaining to various areas of practices in law, from Findlaw~\cite{hughes1999rules} by web crawling. Due to this reason, the system tends to be generalized well over many aspects of law.





\subsection{Text Lemmatization}
\label{Mathod:Lemma}



The linguistic process of mapping inflected forms of a word to the word's core lemma is called lemmatization~\cite{korenius2004stemming}. The crawled natural language text would contain words of all inflected forms. However, the default word2vec model does not run a lemmatizer before the word vector embeddings are calculated; which results in each of the inflected forms of a single word ending up with a separate embedding vector. This in turn leads to many drawbacks and inefficiencies. Maintaining a separate vector for each inflected form of each word makes the model bloat up an consume memory unnecessarily. This especially leads to problems in building the model. Further, having separate vector for each inflected form weakens the model because the values for words originating from the same lemma will be distributed over those multiple vectors. For example, when we search for similar words to the noun input "judge", we get the following words as similar words: \textit{Judge}, \textit{judges}, \textit{Judges}. Similarly, for verb input "train", the model would return the following set of words: \textit{training}, \textit{trains}, \textit{trained}. 

As shown in~\figurename~\ref{fig:Overall methodology}, we use the raw legal text to train the \wvLR~model. But for both \wvLL~model and \wvLLS~model, we first lemmatize the legal document corpus to map all inflected forms of the words to their respective lemmas. For this task we use the Stanford CoreNLP library~\cite{manning2014stanford}.  




\subsection{Training word2vec models}
\label{Method:word2vec}
In this stage we trained wod2vec models. One was on the raw legal text corpus and the other was on the lemmatized law text corpus. Each input legal text corpus included text from over $\CaseCount$ legal case documents amounting to $\WordBillionCount$ billion words in all. The training took over 2 days. As mentioned in the Section~\ref{Mathod:Lemma}, the model trained by the raw legal text corpus is the \wvLR model shown in~\figurename~\ref{fig:Overall methodology}. The model trained on lemmatized law text corpus is \wvLL. Further, a clone of the trained \wvLL is passed forward to~\ref{Method:SemSim} in order to build the \wvLLS model. Following are the important parameters we specified in training these models.

\begin{itemize}
  \item size (dimensionality): 200 
  \item context window size: 10
  \item learning model: CBOW
  \item min-count: 5
  \item training algorithm: hierarchical softmax
\end{itemize}

\begin{figure}[!hb]	
	\centering
\includegraphics[width=0.35\textwidth]{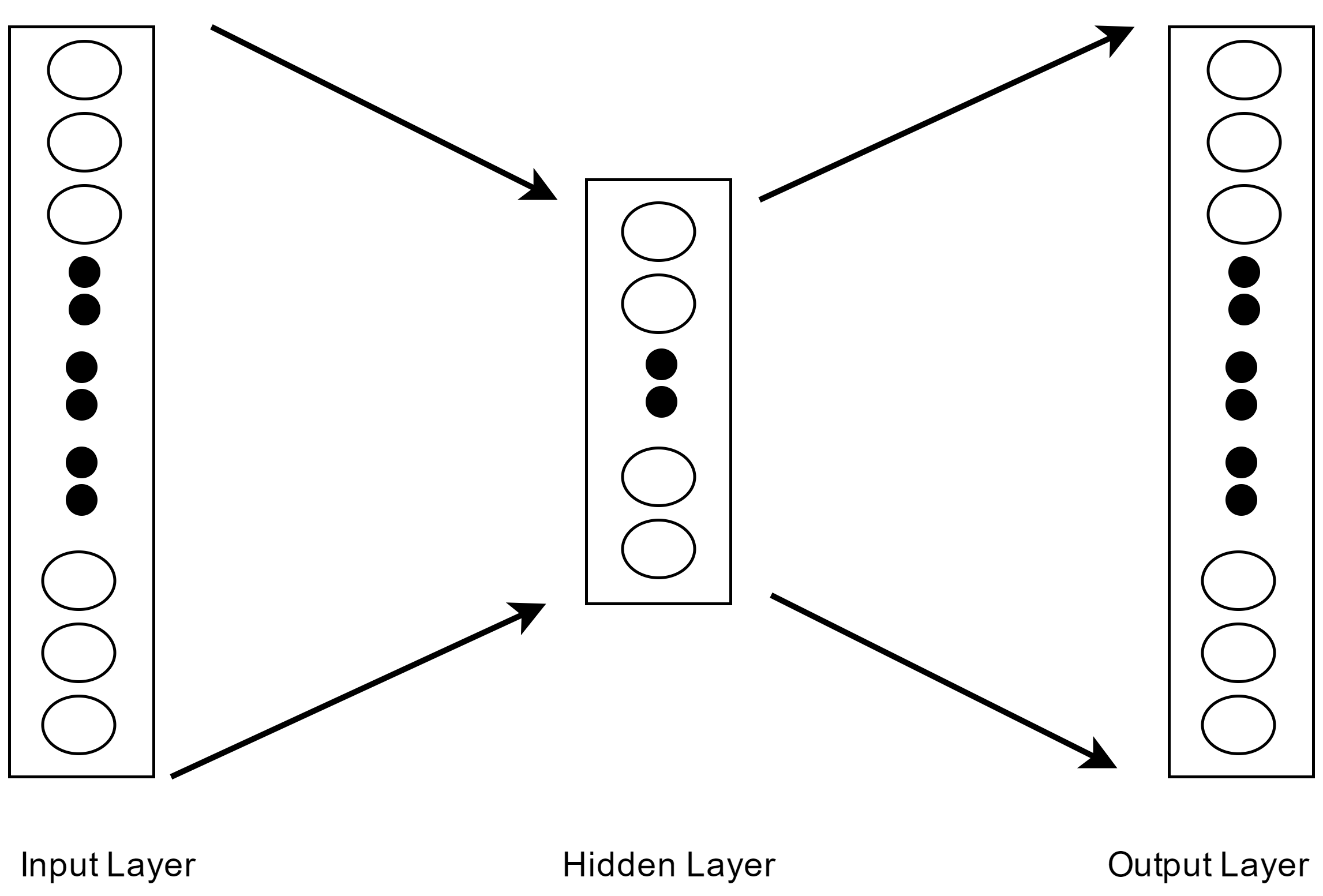}
	\caption{word2vec neural network model: This is a neural network model with a single hidden layer, which is used in the word2vec training process. Input layer and output layer both consist of v neurons where v is the number of words in the vocabulary of the given raw text corpus. The hidden input layer consists of n neurons where n is the intended dimensionality of the vector representation of each word.}	
	\label{fig:word2vec neural network model}
\end{figure}

For this phase, we used a neural network with a single hidden layer as depicted in figure~\ref{fig:word2vec neural network model}. As mentioned above, to train the system to output a user-specified dimensional vector space, we picked the Continuous Bag Of Words approach for learning weights. The rationale as to why CBOW learning model was picked over skip-gram is that, Skip-gram is said to be accurate for infrequent words whereas CBOW is faster by a factor of window size which is also more appropriate for larger text corpora.

In CBOW, the current word or the target word is being predicted based on the context words, within the specified context window size.



In addition to the above trained models, we obtained the \wvG model which was trained by Google on the Google News dataset. As shown in~\figurename~\ref{fig:Overall methodology}, this is a generic text corpus that contain data pertaining to a large number of topics. It is not fine tuned to the legal domain as the three models (\wvLR, \wvLL, and \wvLLS) that we train. However, Google's model was trained over on around 100 billion words that add up to 3,000,000 unique phrases. This model was built with layer size of 300. Due to the general nature and the massive case of \wvG model, it is possible to use the comparison of results obtained by our models against the results obtained by this model to showcase the effectiveness of a model trained using a specific domain in the applications of that domain over a model trained using a generic domain in application on the same specific domain. 

\subsection{Lexical Semantic Similarity Enhancements}
\label{Method:SemSim}
At this step we used established lexical semantic similarity measures to enhance the output. As mentioned in Section~\ref{Method:word2vec}, we get a clone of the trained \wvLL model to use in this process. Unlike in the cases of the previous models where the training system is internal to the word2vec model, here it was important that the model is trained explicitly using n-fold cross validation. As such, a training dataset where each entry is a key value pair is obtained. The key $k$ is a lemma of a word in the legal domain and the value is an array of lemmas of words $G$ that are most relevant in the legal domain to the word lemma used as the key. The said array is of length $l$. Following paragraphs explain how the training of the model and the testing happen for the first fold in n-fold cross validation. It is imperative to understand that the same process will be done for each of the folds subsequently.  

\subsubsection{Mathematical model}
\label{Method:Sem:Math}
The first step was to obtain an entry from the training set and to query the \wvLL model with the key lemma. Let us define an integer $n$ such that $n=Cl$ where $C>1$. When executing the query, \wvLL model was instructed to return the first $n$ elements that matches the query best along with the word2vec similarity values. Let us call this resultant Matrix $R$. $R$ has $n$ columns where $w_i$ is the word similar to $k$ and $d_i$ is the word2vec similarity between $k$ and $w_i$. Equation~\ref{Eq:Rmatrix} shows $R$. 


\begin{equation}
R = 
\begin{bmatrix}
   w_1 & w_2 & w_3 & \dots & w_n \\
   d_1 & d_2 & d_3 & \dots & d_n
\end{bmatrix}
\label{Eq:Rmatrix}
\end{equation}

From $R$, we created the vectors $W$ and $D$ as shown in Equation~\ref{Eq:Words} and Equation~\ref{Eq:word2vecVals} respectively. Each $w_i$ and $d_i$ has the same value as they had in $R$.

\begin{equation}
W = \{ w_1, w_2, w_3,..., w_n \}
\label{Eq:Words}
\end{equation}

\begin{equation}
D = \{ d_1, d_2, d_3,..., d_n \}
\label{Eq:word2vecVals}
\end{equation}

We defined the following functions for words $w_i$ and $w_j$. The lexical semantic similarity calculated between $w_i$ and $w_j$ using the  Wu \& Palmer's model~\cite{Wu1994} was given by $wup(w_i,w_j)$. The lexical semantic similarity calculated by Jiang \& Conrath's model~\cite{Jiang1997} was given by $jcn(w_i,w_j)$. $hso(w_i,w_j)$ was used to indicate the lexical semantic similarity calculated using the Hirst \& St-Onge's system~\cite{hirst1998lexical}.

Next we created the lexical semantic similarity matrix $M$. The matrix $M$ is a $4 \times N$ matrix where $N$ has the same value as in Equation~\ref{Eq:Rmatrix}. The first row element $m_{1,i}$ of $M$ was calculated by taking the  Wu \& Palmer similarity between $k$ and $w_i$. Here $w_i$ element at index $i$ of $W$. Thus, $m_{1,i}$ is equal to $wup(k,w_i)$. Similarly, the second row element $m_{2,i}$ of $M$ was calculated by taking the  Jiang \& Conrath similarity between $k$ and $w_i$. The third row consists of Hirst \& St-Onge similarities while the forth row contains a series of $1$s for the sake of the bias. The matrix $M$ is shown in Equation~\ref{Eq:SemSim}.

\begin{equation}
M = 
\begin{bmatrix}
   wup(k,w_1) & wup(k,w_2) & \dots & wup(k,w_n) \\
   jcn(k,w_1) & jcn(k,w_2) & \dots & jcn(k,w_n) \\
   hso(k,w_1) & hso(k,w_2) & \dots & hso(k,w_n) \\
   1 & 1 & \dots & 1
\end{bmatrix}
\label{Eq:SemSim}
\end{equation}

We defined the normalizing function given in Equation~\ref{Eq:norm} to return a value $x_{norm}$ when given a value $x_{raw}$ that exists between the maximum value $v_{max}$ and the minimum value $v_{min}$. The returned $x_{norm}$ is a double value that has the range $[0,1]$.

\begin{equation}
x_{norm} = normalize(x_{raw},v_{min},v_{max})
\label{Eq:norm}
\end{equation}

We created the matrix $SM$ from the matrix $M$ by calculating each $sm_{i,j}$ using the Equation~\ref{Eq:norm} on each $m_{i,j}$. For this we used the $min$ and $max$ values shown in table~\ref{tab:SemSimMinMax}. 

\begin{table}[!htb]
	\centering
	\caption{Lexical Semantic Similarity Statistics}\label{tab:SemSimMinMax}
	\begin{tabular}{|l|c|c|}
		\hline
        \textbf{Model} & \textbf{Min} & \textbf{Max} \\
		\hline
		Wu \& Palmer & 0.0 & 1.0\\
		\hline
		Jiang \& Conrath & 0.0 & $\infty$ \\
        \hline
		Hirst \& St-Onge & 0.0 & 16.0\\
        \hline
	\end{tabular}	
\end{table} 

We defined the value matrix $V$ using the vector $D$ and the Matrix $SM$ as shown by Equation~\ref{Eq:value}.  

\begin{equation}
V = 
\begin{bmatrix}
  D \\
   SM
\end{bmatrix}^{T}
\label{Eq:value}
\end{equation}

We defined the vector $E$ where each element $e_i$ is given by activating a neural network by values $v_{i}$, where $v_{i}$ is the $i$th row of $V$. The training of the aforementioned neural network is explained in Section~\ref{Method:Sem:ml}.  



\subsubsection{Machine Learning for weight calculation}
\label{Method:Sem:ml}
The motive behind using machine learning is to train the system to take into account how each similarity measure value can be used to derive a new, compound, and better representative value for similarity. As mentioned in Section~\ref{Method:Sem:Math}, a neural network was chosen as the machine learning method. Initially the weight values were initialized to random values and $E$ was calculated. Next the matrix $MI$ was defined as shown in Equation~\ref{Eq:InterMediate}.

\begin{equation}
MI = 
\begin{bmatrix}
  W \\
   E
\end{bmatrix}
\label{Eq:InterMediate}
\end{equation}

The matrix $Y$ was obtained by sorting the columns of matrix $MI$ in the descending order of elements in the second row. We defined a seeking function given in Equation~\ref{Eq:seek} to return the index of word $w$ in the first row of matrix $P$. If the word $w$ does not exist in the first row of matrix $P$, it returns the column count of the matrix $P$. Also, observe that the new matrix $Y$ has the same form as the initial matrix $R$ shown in Equation~\ref{Eq:Rmatrix}. This symmetrical representation is important because it gives us the opportunity to use the same accuracy measures on all the word2vec models in Section~\ref{Results} to achieve a fair comparison.

\begin{equation}
s = seek(w,P)
\label{Eq:seek}
\end{equation}

Next we defined the error $err$ according to Equation~\ref{Eq:err} where the value of $x_i$ was derived from Equation~\ref{Eq:xi} and $\epsilon$ is a small constant.

\begin{equation}
x_i = 
\begin{cases}
    1,& \text{if } seek(g_i,Y) < l \\
    \frac{n-seek(g_i,Y)}{n-l},              & \text{otherwise}
\end{cases}
\label{Eq:xi}
\end{equation}

\begin{equation}
err = 1 - \frac{\epsilon+\displaystyle \sum_{i=1}^{l} x_i }{|G \bigcap W|+\epsilon} 
\label{Eq:err}
\end{equation}


This error $err$ is used to adjust the neural network. This training cycle is continued until convergence. The completed model that is trained this way is named \wvLLS model. 

\subsection{Query processing}
In order to use and test our models, we built a query processing system. A user can enter a query in the legal domain using the provided interface. The system then takes the query and applies natural language processing techniques such as PoS tagging until the query is through the NLP pipeline to be lemmatized. We used the same Stanford Core NLP pipeline that we used in Section~\ref{Mathod:Lemma} for this task. The reason for this is to bring all the models to the same level to be compared equally. We have shown this step in~\figurename~\ref{fig:Overall methodology}.


\subsection{Experiments}

We got experts in the legal field to create a golden standard to test our models. The golden standard includes $\GoldenConceptCount$ concepts with each containing $\GoldenLength$ words that are most related to the given concept in the legal domain picked from a pool of over $\GoldenPool$ words by the legal experts. 



The accuracy levels of these experiments are measured in terms of precision and recall~\cite{wimalasuriya2010ontology}, which are common place measures in information retrieval. The logical functionality of these two are based on the comparison of an expected result and the effective result of the evaluated system.


If the Golden Standard Word Vector is $G$ and the word vector returned by the model is $W$ (Same naming conventions as Section~\ref{Method:Sem:Math}), the recall in this study is calculated with equation~\ref{Eq:recall}. This measures the completeness of the model. Our recall calculation used the same function suggested in~\cite{wimalasuriya2010ontology}. 

\begin{equation}
recall = \frac{|G \bigcap W|}{|G|} 
\label{Eq:recall}
\end{equation}

The precision calculation in this study is not as clear cut as it is described in~\cite{wimalasuriya2010ontology}. This is because in those systems the precision is only a matter of set membership and thus would simply be ratio between the correctly found similar words over the total number of returned words. However, in the case of word2vec models, it is the user who input the number of matching words to retrieve. Thus, it is wrong to use the total number of returned words to calculate the precision in cases where there is prefect recall. In the cases of imperfect recall, the classical precision is adequate.

While word2vec make precision calculation difficult as shown above, it also has a quality that makes finding the solution to that problem somewhat easy. That is the fact that the returning word list of similar words is sorted in the descending order. This is the same property that we used in the above Section~\ref{Method:Sem:ml} to calculate the error. Thus it is logical to derive that the precision is given by equation~\ref{Eq:precision} where $err$ is the error calculated by equation~\ref{Eq:err}. 


\begin{equation}
precision = 1-err 
\label{Eq:precision}
\end{equation}







\begin{table*}[!htb]
	\centering
	\caption{Results comparison (P=Precision, R=Recall)}\label{tab:ResComp}
	\begin{tabular}{|l|c|c|c|c|c|c|c|c|c|c|c|c|}
		\hline
		 \multirow{2}{*}{Model} & \multicolumn{3}{c}{k=20} \vline & \multicolumn{3}{c}{k=50} \vline & \multicolumn{3}{c}{k=100} \vline & \multicolumn{3}{c}{k=200} \vline  \\ 
		\hhline{~------------}
        & P & R & F1 & P & R & F1 & P & R & F1 & P & R & F1 \\
        \hline
		\wvG & 0.57 & 0.19 & 0.29 & 0.62 & \textbf{0.33} & 0.43 & 0.67 & 0.41 & 0.51 & 0.74 & 0.46 & 0.57 \\
		\hline
		\wvLR & \textbf{0.75} & 0.19 & 0.30 & 0.71 & 0.31 & 0.43 & 0.74 & 0.38 & 0.51 & \textbf{0.77} & 0.44 & 0.56 \\
        \hline
		\wvLL & 0.73 & 0.22 & 0.34 & 0.72 & 0.32 & 0.45 & \textbf{0.75} & 0.40 & 0.52 & 0.76 & 0.47 & 0.58 \\
        \hline
		\wvLLS & 0.66 & \textbf{0.24} & \textbf{0.36} & \textbf{0.73} & \textbf{0.33} & \textbf{0.52} & 0.72 & \textbf{0.43} & \textbf{0.54} & 0.74 & \textbf{0.50} & \textbf{0.60} \\
		\hline
	\end{tabular}	
\end{table*} 


\section{Results}
\label{Results}

This section includes results obtained from the four different models (\wvG, \wvLR, \wvLL, and \wvLLS) as introduced in Section~\ref{Method:word2vec}. Results shown in table~\ref{tab:ResComp} were obtained for different $k$ values, where $k$ is the number of words requested from each of the models. As expected, the F1 of each model increases with $k$, where the possibility of finding the correct similar words against the golden standard increases.  Given that the task here is to return the expected set of words, the recall is more important than precision (i.e: False-Negatives are more harmful than False-Positives). In that light, it is obvious that the \wvLLS performs better than all other models because it consistently has the highest recall for all values of $k$. In addition to that, the \wvLLS model also has the highest $F1$ for all values of $k$, which is sufficient proof that the small loss in precision does not adversely affect the overall result.


\begin{figure}[!hb]	
	\centering
\includegraphics[width=0.5\textwidth]{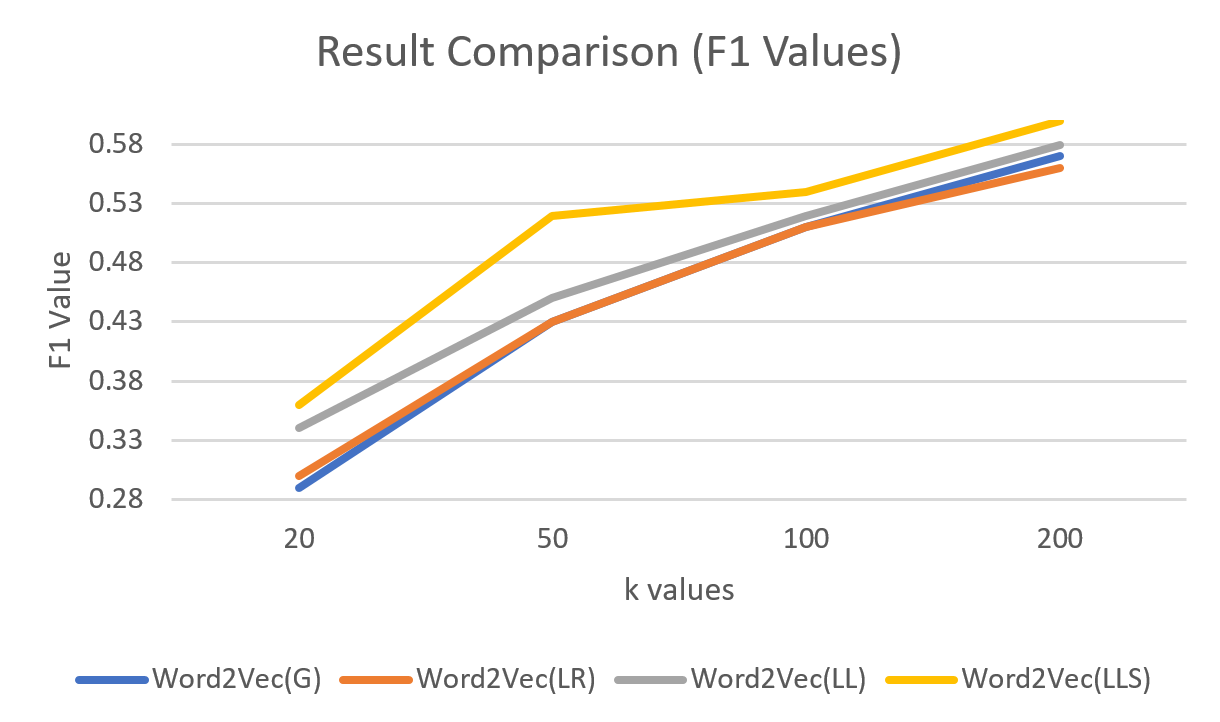}
	\caption{F1 value comparison}	
	\label{fig:F1 value comparison}
\end{figure} 

A graphical comparison of the changes in the $F1$ measure is shown in~\figurename~\ref{fig:F1 value comparison}. As shown, the domain specific models such as \wvLR, \wvLL, and \wvLLS, show better results than the general \wvG model. It should be noted that this performance enhancement has happened despite the fact that \wvG was trained on a text corpus 3 times bigger than the text corpora we have used in this study for the models \wvLR, \wvLL, and \wvLLS.


In the comparison of domain specific models, we can see a clear distinction between the \wvLR~and \wvLL~models, where the \wvLL~generally performs better. Further, it can be observed that the \wvLLS~model outperforms both \wvLR~and \wvLL~models.   



\section{Conclusion and Future Works}
\label{conclusion}
The hypothesis of this study had three main claims and each of these claims were justified by the results presented in Section~\ref{Results}. The first claim of the hypothesis is that a word embedding model trained on a small domain specific corpus can out perform a word embedding model trained on a large but generic corpus. The success of \wvLR model over the \wvG model justifies this claim. In Section~\ref{Mathod:Lemma} we proposed the second claim: word lemmatization, which removes inflected forms of words and improve the performance of a word embedding model. \wvLL model obtaining better results than \wvLR model proved this claim to be true. The third claim was made in Section~\ref{Method:SemSim}. There, we proposed that usage of lexical semantic similarity measures trained over a machine learning system can improve the overall system performance. The significant improvement that we show for the \wvLLS model over the \wvLL model verified this claim also to be accurate. Therefore we can conclude that the proposed methodology of word vector embedding augmented by lexical semantic similarity measures, gives a more accurate evaluation of the extent of which a given pair of words is semantically similar in the considered domain.



Semantic similarity measures are important in many areas of applications. Out of those, for future work, we expect to extend the findings of this study to the document level. Word based semantic similarity is the building block for sentence similarity measures, which in turn aggregates to build document similarity measures. This is the direction in which we intend to move. We will be using this word semantic similarity measure to build up to a document similarity measure which can be used for more efficient domain based document retrieval systems.  

\bibliographystyle{IEEEtran}
\bibliography{word2vec}

\end{document}